\documentclass[runningheads]{llncs}

\usepackage[utf8]{inputenc}
\usepackage[english]{babel}
\usepackage[backend=bibtex,style=numeric]{biblatex}  
\usepackage{csquotes}
\usepackage{amsmath}
\usepackage{graphicx}
\usepackage{graphics}
\usepackage{moreverb}
\usepackage{diagbox}
\usepackage[colorlinks=true, citecolor=blue,linkcolor=blue,bookmarks=false]{hyperref}
\usepackage{epsfig}
\usepackage{txfonts}

\usepackage{newclude}
\usepackage{float}
\usepackage{multirow}  
\usepackage[export]{adjustbox}
 \usepackage[labelsep=quad,indention=10pt]{subfig}
\PassOptionsToPackage{subfigure}{tocloft}
\usepackage{fixmetodonotes}

\renewcommand{\caption}[1]{\singlespacing\hangcaption{#1}\normalspacing}

\title{Adversarial Attacks on Convolutional Neural Networks in Facial Recognition Domain}
\titlerunning{Adversarial Attacks on Convolutional Neural Networks in Facial Recognition Domain}

\begin{document}

\maketitle

\begin{center}
\normalsize
Yigit Alparslan$^{1}$, Ken Alparslan$^{2}$, Jeremy Keim-Shenk$^{1}$, Shweta Khade$^{1}$, Rachel Greenstadt$^{3}$\\

$^{1}$ Department of Computer Science, Drexel University, Philadelphia, PA, US, 19104\\
$^{2}$ Department of Computer Science, Conestoga College, Waterloo, ON, CA \\
$^{3}$ Computer Science Department, New York University, New York, NY, US, 10012

ya332@drexel.edu, kalparslan6724@conestogac.on.ca, jdk346@drexel.edu, sk3783@drexel.edu, greenstadt@nyu.edu

\end{center}

\begin{abstract}
Numerous recent studies have demonstrated how Deep Neural Network (DNN) classifiers can be fooled by adversarial examples, in which an attacker adds perturbations to an original sample, causing the classifier to misclassify the sample [6]. Adversarial attacks that render DNNs vulnerable in real life represent a serious threat, given the consequences of improperly functioning autonomous vehicles, malware filters, or biometric authentication systems [10]. In this paper, we apply Fast Gradient Sign Method to introduce perturbations to a facial image dataset and then test the output on a different classifier that we trained ourselves, to analyze transferability of this method. Next, we craft a variety of different attack algorithms on a facial image dataset, with the intention of developing untargeted black-box approaches assuming minimal adversarial knowledge, to further assess the robustness of DNNs in the facial recognition realm. While experimenting with different image distortion techniques for the purpose of determining weaknesses of the DNNs, we focus on modifying single optimal pixels by a large amount, or modifying all pixels by a smaller amount, or combining these two attack approaches. While our single-pixel attacks achieved about a 15\% average decrease in classifier confidence level for the actual class, the all-pixel attacks were more successful and achieved up to an 84\% average decrease in confidence, along with an 81.6\% misclassification rate, in the case of the attack that we tested with the highest levels of perturbation. Even with these high levels of perturbation, the face images remained fairly clearly identifiable to a human. Understanding how these noised and perturbed images baffle the classification algorithms can yield valuable advances in the training of DNNs against defense-aware adversarial attacks, as well as adaptive noise reduction techniques [6]. We hope our research may help to advance the study of adversarial attacks on DNNs and defensive mechanisms to counteract them, particularly in the facial recognition domain.

\end{abstract}

\section{Introduction}

Recognizing the identity of an image of a human face is a relatively easy job for humans in most cases, assuming that we have seen the person’s face before. However, these types of image recognition tasks represent complex challenges for computers, requiring advanced classification models that process high-dimensional data inputs. In recent decades, advances in deep neural networks (DNN) have allowed computers to achieve or even exceed human-level performance on difficult image recognition tasks. DNNs are widely used today in several critical fields, such as bio-authentication systems, autonomous vehicles, malware detection, and spam filtering. Facial recognition is one of these fields where convolutional DNNs are frequently used and generally very effective. Nevertheless, there are several challenges that accompany the benefits of use of DNNs. One problem is that DNN classifiers can be fooled by adversarial examples, which are crafted by adding perturbations into an original sample [6].

Facial recognition represents an interesting and important area for studying adversarial attacks on DNN classifiers. On one hand, some users of social networks like Facebook and Instagram may be interested in being able to apply filters (i.e. perturbations) to the photos they upload that disrupt these social network services’ identity recognition algorithms, for reasons of privacy (e.g. preventing tag suggestions) or resisting state surveillance in oppressive regimes. On the other hand, social network companies, law enforcement, and various commercial interests may wish to be able to use image classification tools with defensive mechanisms that are robust to adversarial attacks, in order to reduce these attacks’ effectiveness. Wide usage of DNNs makes the problem of creating robust and secure DNNs even more important in safety-critical applications. We are interested in exploring whether there are any particular characteristics of facial recognition which seem to make adversarial attacks more or less successful in this area.

Current attacks that have been studied in the field, such as those of Carlini [6] and Papernot et al. [10] have been studied as proof-of-concepts, where adversarial attackers are assumed to have full knowledge of the classifier (e.g. model, architecture, model weights, parameters, training and testing datasets). The strongest attack in the literature at the time of

writing this article is Carlini’s attack based on the L2 norm, and it is a white-box attack requiring full knowledge of the model. Much of this research has been interested in developing the most effective attacks possible, to be used as standards against which to test the robustness of image classifier DNNs. With less knowledge of the classifier model, the effectiveness of the attack decreases.

There is also interest in crafting attacks that assume minimal knowledge of the adversary regarding the classifier model, since in most real world applications the adversary does not have access to the classifier’s parameters unless the adversary is an insider. Attacks that have been studied in the
current literature \cite{cw17} \cite{gss15} \cite{llslsw} \cite{mff16} \cite{pmwjs16} \cite{pmjfcs16}  have typically used the MNIST and/or CIFAR-10 datasets as a proof-of-concept. Such standard datasets provide a common ground for different research labs all around the world. However, there is an interest in using facial image datasets while crafting attacks. Use of face images in our studies can help us shed light into implications for current industry application areas where facial recognition is a central focus, such as tagging in social media, surveillance, privacy, and bio-authentication systems. In order to study neural networks in the facial recognition field instead of with standard datasets such as MNIST or CIFAR-10, we experimented with different sets of procedures to craft adversarial examples on a facial dataset and see how robust a facial classifier was to these attacks.

We were able to craft adversarial attacks to face images which achieved up to an 81.6\% misclassification rate against the classifier that we trained, using an approach that altered all pixels in the input image with alternating additions and subtractions to pixel values, while assuming nothing about the classifier. The perturbed images produced by this attack, while visibly altered, were still clearly identifiable to a human. In this paper, we discuss the different approaches to attacks that we developed, including altering single optimally chosen pixels, altering all pixels, or some combination of these approaches, and we compare results across these attacks.

\section{Background}
We do not yet fully understand how the brain solves visual object recognition. However, in the last few years, the field of machine learning has made tremendous progress on developing models that mimic what the brain achieves. In particular, deep neural networks (DNN) using convolutional structures have achieved reasonable performance on hard visual recognition tasks. In convolutional neural networks, subsets of input attributes (in our case, pixel values) representing regions of images are together connected to nodes in subsequent layers of the neural network. These convolutional architectures allow the neural network to learn to extract “features” from the images to reduce dimensionality and focus on characteristics of the images that are useful for classification purposes.

Input vectors to DNNs for image recognition represent matrices of values for each pixel in an image. In our study, we used a grayscale image dataset, so each pixel’s representation was simply a single integer between 0 (black) and 255 (white) indicating how dark each pixel is. In the case of color images, on the other hand, each pixel typically has three values for the three color channels red, green, and blue. The perturbations that we applied to images consisted of adding or subtracting from the integer values representing each pixel. As described in the following sections, altering just one or two strategically placed pixels can reduce face classifier confidence levels substantially, however the task of searching for an optimal pixel, while not assuming any knowledge of model architecture or weights (only output confidence levels), is computationally highly time-expensive, given that there are typically thousands of pixels to test even for relatively small, low-resolution images, and each pixel modification requires another call of the classifier with the modified image as input.

\section{Related Work}
Papernot, et al. \cite{pmjfcs16} studied the limitations of Deep Neural Networks when attacked by adversarial attacks. Their algorithms could reliably produce samples correctly classified by human subjects but misclassified in specific targets by a DNN with a 97\% adversarial success rate, while only modifying on average 4.02\% of the input features per sample.

Then, in 2015, the same laboratory group devised a new defense technique called Defensive Distillation to prevent adversarial attacks \cite{pmwjs16}. Their “distillation” technique leads gradients used in adversarial sample creation to be reduced by a factor of 1030. Their study showed that defensive distillation can reduce effectiveness of adversarial sample creation from 95\% to less than 0.5\% on a studied DNN \cite{pmwjs16}. 
In 2016, Carlini and Wagner \cite{cw17} tested three new attacking algorithms (CW attacks) and found that their algorithm tailored around the “L2” norm was successful with 100\% probability even against distilled neural networks. 
Much of the related literature that we have reviewed, including the work of Carlini and Wagner \cite{cw17}, have used CIFAR-10, MNIST, or ImageNET datasets. In order to further advance this field of study in a realm with major practical applications, we decided to apply adversarial techniques to the facial recognition domain, to explore how these attacks translate to this domain.
Different attack algorithms have different approaches in terms of how much they assume the attacker knows about the classification DNN structure and weights, or about the training and testing datasets used. Knowing more about the classifier and datasets allows the attacker to develop more fine-grained perturbations that require less substantial modifications to achieve the same desired misclassification effect. Such situations may be less realistic in most real world application areas, although insider threats could theoretically represent a security issue in some cases (e.g. a Facebook employee who releases information about how to circumvent their tagging recognition algorithms). In our work, we focused on attack algorithms that assumed relatively little knowledge about the DNN classifier, taking a comparably more black-box approach to crafting attacks.

\section{Approach}

Our first step was to choose a face image dataset. We decided to train an image classifier on the Extended Yale Face Database B cropped image database, which contains 2470 images of 38 individuals (65 images per individual) in various lighting conditions \cite{gbk01} \cite{lhk05}. Each image in this dataset is 168 x 192 pixels and in grayscale, and we converted them into JPEG format (from PGM) for compatibility reasons with the classifier model that we trained. We divided the database into two sets: a training set and a testing set. A center light pose was selected for each individual for our testing dataset. These images represented cases which the classifier had a relatively easy time classifying, compared to some of the other images in the dataset that were more shadowed. The testing to training data set ratio was 1 to 64.

Next, we trained a neural network using Google’s publicly available Inception v3 convolutional neural network model architecture \cite{svisw16} on this Yale B dataset. We used a feature extraction module in which the earlier layers of the neural network are first trained on the ImageNet database to learn to generate useful features for classification purposes from raw image data, and then subsequent layers are trained to produce classifications based on the features extracted for the specific training dataset provided by the \cite{tftutorial}.

Training was completed in 35 minutes on a standard Google Cloud Compute Engine \cite{c18}. The publicly available open-source framework TensorFlow in Python was used in the training and testing process. The training and cross-validation process was completed over 5000 steps with a final test accuracy of 82.5\% on the training dataset. The classifier achieved 100\% accuracy in classifying the images in the separate testing dataset that we used for generating attacks, before altering these images.

After training the DNN, we adapted the current literature for our study by using Fast Gradient Sign Method to create adversarial examples with our dataset, but with the attacks based off of a classifier model from the study of Papernot, et al. \cite{pmwjs16} with identical parameters as in the study. Fast Gradient Sign Method assumes full knowledge of the image classifier model, including the confidence level of classification results, model weights, and parameters. The attack algorithm seeks to maximize the loss function of the classifier by adding perturbations in the direction of the loss gradient. We then tested the perturbed images generated through this approach against our own model to analyze transferability, and found that they had a relatively small impact on reducing our classifier’s confidence compared to other attacks that we developed. This was not particularly surprising, given that the perturbations were generated for a different model, by using a white-box approach closely tailored for that other model.

After the adaptation of this attacking algorithm with our facial image database, we were interested in crafting adversarial examples with less assumptions about the adversary’s knowledge of the classifier they are attacking. Therefore, we created our own attack algorithms in more of a black-box setting. The attacks that we developed in our study are summarized in the following sections.

\subsubsection{Approach 1: $L_0$ attacks}
To begin, we were interested in seeing whether there were certain pixels in the face images which, if altered, would cause a relatively greater change in the classifier’s confidence levels for the correct class for each image, compared to other pixels.

In this approach, the adversary was assumed to be searching for an attack with the limit $\Vert L_0 \Vert <= 2$. In other words, the adversary could change only at most 2 pixels on an image.  

For each testing image (168 x 192 pixels), we divided the image into 56 squares of 24 x 24 pixels. From each of the 56 squares, one at a time, we randomly selected a pixel and altered its value by 128, mod 255 (“inverting” the pixel value, in a sense, since each pixel’s value is in the range from 0 to 255), and then ran the classifier on the resulting image. After testing each of 56 test pixels, we selected the one (in Attack A: $L_0$ distance = 1) or two (in Attack B: $L_0$ distance = 2) pixels that resulted in the greatest decrease in the classifier’s confidence that the image was the actual class. The image resulting from changing these one or two pixels was chosen as the attack image. 

\begin{figure}[htbp!]
\centering
  \includegraphics[width=\linewidth]{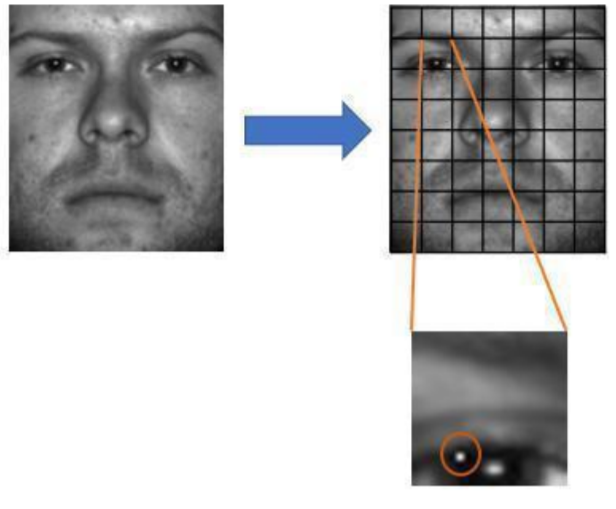}
  \caption{Visualization of attack based on approach 1. Approach 1 involved randomly selecting a single pixel from each of 56 squares overlaid on the input image and testing the result of inverting this pixel’s value on the classifier.}
  \label{fig:approach1}
\end{figure}

In Attack C, we used this same concept, but in an iterative manner, somewhat influenced by the Fast Gradient Sign Method. This attack used a greedy approach by first selecting the best pixel (the pixel that would result in the most confidence decrease) from the 56 squares and saving the resulting image from this one pixel change. It then completed a second iteration of the individual pixel testing process, again randomly choosing another set of pixels from the 56 squares and testing the effects of changing each of those pixels of the already modified image on the classifier’s performance. $L_0$ distance was equal to 1 for this attack.

A downside of this approach was that it was highly time-expensive to repeatedly modify and save the images and then classify the resulting images. The Attack C iterative approach (the slowest one, since it required classifying 112 modified images per original image) took a few hours to complete attack testing for all of the 38 images in the testing dataset.

Summary of specific attacks developed using Approach 1:

\begin{enumerate}
    \item Attack A: Change 1 Best Pixel’s Value by 128 (Test 56). $\Vert L_0 \Vert $ = 0
 \item Attack B: Change 2 Best Pixels’ Values by 128 (Test 56). $\Vert L_0 \Vert $ = 2
 \item Attack C: Change 1 Best Pixel’s Value by 128 - 2 Iterations (Test 56 + 56). $\Vert L_0 \Vert $ = 1
\end{enumerate}

\subsubsection{Approach 2: $L_2$ attacks}

In this approach, the adversary was assumed to be searching for an attack with no limit on $\Vert L_0 \Vert$. In other words, the adversary could change any number of pixels. Additionally, the total amount of change applied to each pixel would determine the $L_2$ amount for that attack.

In this set of attacks, we wanted to analyze the effect on the classifier’s confidence if we altered all of the pixels in the testing images. We considered this to be a more realistic real-world attack, since, unlike Approach 1, it does not necessarily depend on the ability to test confidence levels returned by the classifier for each image and class after introducing perturbations. After determining an effective magnitude of perturbation to introduce in general (based on empirical results in testing against a particular classifier), one could simply apply this level of perturbation against any classifier with the hopes that it could be relatively effective in causing misclassifications. Our task was to determine what levels of perturbation produced
high rates of misclassification for our classifier, while producing images that were still identifiable to a human, and how large these rates of misclassification were in comparison to the previous more individual pixel-oriented approach.

We began by altering all of the pixels in the face images by a small value in the same direction (i.e. making them all darker or all lighter by some small magnitude). We then iterated with increasing levels of perturbation until the resulting images were
misclassified by the classifier. 
We found that when the pixel values were all modified in the same direction, by the time they were misclassified, the resulting images were often either whitewashed (in the case of addition) or shadowed (in the case of subtraction) to the point where they had become unrecognizable to a human. We then thought of instead taking a checkerboard-like approach, where we alternated between adding and subtracting values from each pixel. This resulted in a grainy or pixelated image, but most of the images were still identifiable
to a human by the time that they were altered enough to fool the classifier. 

At this point, we experimented with different magnitudes of perturbation (different $L_2$ amounts) to try to find an amount which resulted in frequent
misclassification while still producing images that were identifiable to a human and not blatantly clearly altered. Through an iterative testing process with increasing levels of perturbation, we found that about half of the images were misclassified after altering all pixels’ values by 30 or less, while others required larger magnitudes of changes. Five out of 38 of
the testing images were still not misclassified after altering all pixels’ values by over 120. The few images that were particularly difficult to get the classifier to misclassify appeared to represent minority groups within our face database, suggesting that if the classifier had been trained on a larger and more diverse training dataset, then it perhaps could have been more easily fooled in cases like these.

\begin{figure}[htbp!]
    \centering
  \includegraphics[width=0.7\linewidth]{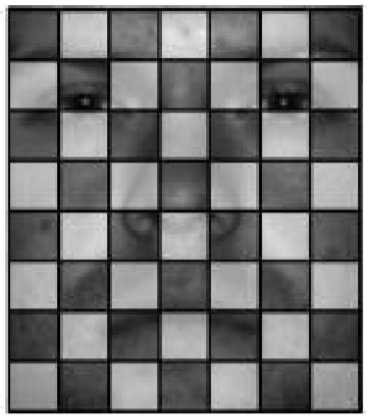}
  \caption{Visual representation of the checkerboard approach. Checkerboard approach included
alternating between adding to and subtracting from pixel values in each of the squares.}
  \label{fig:checkerboard}
\end{figure}

It can intuitively be imagined that adding some relatively small degree of randomization might further contribute to noising the images to decrease classifier confidence in the correct class while keeping them identifiable to a human. Therefore, we decided to randomize the magnitude of change for each pixel
within some range, while still alternating between adding or subtracting for each pixel. We settled upon some ranges of perturbation to use in our final attacks based on the empirical evidence from our iterative tests of increasing levels of perturbation described above. Based on our results, we crafted attacks which chose a randomized magnitude of perturbation for each pixel ($L_2$) either between 30 and 60 Attack D, between 60 and 90 (Attack E), or between 120 and 150  (Attack F). 

Summary of specific attacks developed using Approach 2:
\begin{enumerate}
    \item Attack D: Change All Pixels’ Values by 30 to 60 (Randomly Selected Value), Alternating between
Additions and Subtractions. ($30 < \Vert L_2 \Vert < 60$)
 \item Attack E: Change All Pixels’ Values by 60 to 90
(Randomly Selected Value), Alternating between
Additions and Subtractions. ($60 < \Vert L_2 \Vert < 90$)
 \item Attack F: Change All Pixels’ Values by 120 to 150 (Randomly Selected Value), Alternating between
Additions and Subtractions. ($120 < \Vert L_2 \Vert < 150$)
\end{enumerate}

\subsubsection{Combined Approach: $L_0 + L_2$ Attacks}

For one final experiment, we decided to combine the two
general approaches described above by first performing Attack D (altering all pixels’ values by 30 to 60, 30 $< \Vert L_2 \Vert <$ 60) and then attacking the
resulting image again using Attack B (change best 2 pixels’ values by 128 after testing 56, $\Vert L_2 \Vert$ = 2).

Attack G: Dual Attack
\begin{enumerate}
    \item Change All Pixels’ Values by 30 to 60 (Randomly Selected Value)
    \item Change 2 Best Pixels’ Values by 128 (Test 56)
\end{enumerate}

\section{Evaluation}

For the purposes of this study, we sought to design untargeted attacks, whereby the attacker does not have a specific
target incorrect class that they want the classifier to classify
each image as, but rather simply wants the classifier to
misclassify the image as any incorrect class. Based on this, in
order to test the level of success of our
attacks, we focused on two metrics: 
\begin{enumerate}
    \item  ConfA : the percent decrease in classifier confidence in the actual class for each image after it has been altered, relative to the baseline confidence for the unaltered image 
    \item PercentMisclass: the percent of images from the testing dataset that were misclassified after applying the attack.
\end{enumerate}

More formally,
\begin{equation}
    ConfA = \frac{ConfA(BaselineImg)\times ConfA(AttackImg)}{ConfA(BaselineImg)}
\end{equation}
where ConfA represents the classifier’s confidence that the
image is its actual class.
\begin{equation}
    MissClass = \frac{Result^{B}_{A}}{Result^{A}_{A} + Result^{B}_{A}}
\end{equation}

\begin{figure}[htbp!]
    \centering
  \includegraphics[width=0.7\linewidth]{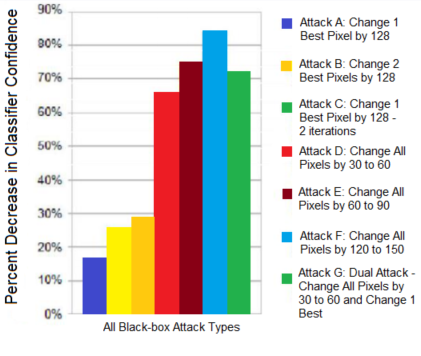}
  \caption{Graph of average percent decrease ($ConfA$) in confidence for actual class after each attack. Attack F results in the most percent decrease. However, the changes are slightly visible to the human eye.}
  \label{fig:results1}
\end{figure}

where $Result^{B}_{A}$ represents the number of images of class A
that were misclassified as some other class B and ResultA
A
represents the number of images of class A that were correctly classified as class A.

\begin{figure}[htbp!]
    \centering
  \includegraphics[width=0.6\linewidth]{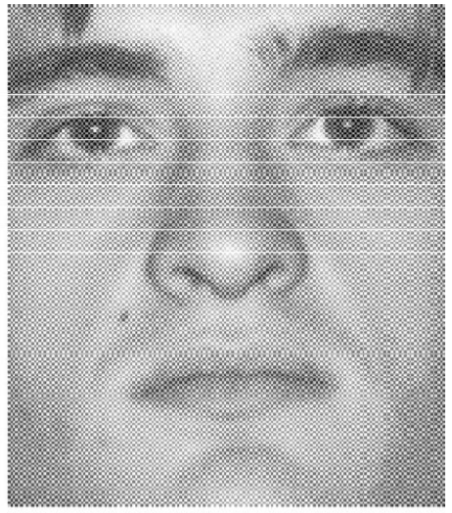}
  \caption{Example of altered image using Attack F. This attacked was created by changing all pixel values with a random pixel amount between 120 and 150.}
  \label{fig:attack_f_example}
\end{figure}

\begin{table}
\centering
\caption{All results from different adversarial attacks. Reduced confidence levels as well as number of samples that have been misclassified are reported for each type of attack. All attacks are done over the testing dataset. Attack type F, which is changing all pixels by some amount delta $\delta$ seem to lower the model accuracy the most compared to changing select pixels by delta $\delta$.} \label{table:all_results_combined}
\vspace{2em}
\scalebox{0.9}{
\begin{tabular}{|p{2cm}|p{6cm}||p{2cm}|p{3cm}|  }
 \hline
 \multicolumn{4}{|c|}{\textbf{Results from Attacks}} \\
 \hline
 \multicolumn{2}{|c|}{\textbf{Attack Type}}& \textbf{Confidence Reduction} & \textbf{\% of Misclassifications}\\
 \hline
& Attack A: Change 1 Best Pixel by 128 (Test 56) & 16.2\% & 0.0\%\\
 \cline{2-4}
\multirow{2}{*}{\textbf{$L_0$ Attacks}} & Attack B: Change 2 Best Pixels by 128 (Test 56) &   25.7\%  & 2.6\%\\
 \cline{2-4}
&Attack C: Change 1 Best Pixel by 128 - 2 Iterations
(Test 56 + 56) &28.7\% & 0.0\%\\
 \cline{1-4}
&Attack D: Change All Pixels by 30 to 60   &65.9\% & 44.7\%\\
 \cline{2-4}
\multirow{1}{*}{\textbf{$L_2$ Attacks}} &Attack E: Change All Pixels by 60 to 90&   74.6\%   & 65.8\%\\
 \cline{2-4}
 &Attack F: Change All Pixels by 120 to 150& \textbf{84.2\%}   & \textbf{81.6\%}\\
  \hline
\multirow{1}{*}{\textbf{$L_0 + L_2$}}&\multirow{2}{*}{Attack G: Dual Attack - 30 to 60}& \multirow{2}{*}{72.0\%}   & \multirow{2}{*}{60.5\%}\\
\textbf{Attacks}& & & \\
 \hline
\end{tabular}
}
\vspace{2em}
\end{table}

A full table of the results we obtained for these two metrics
for each of the seven attacks that we developed can be found
in the \autoref{table:all_results_combined}. A bar graph of the results for ConfA for each
of these attacks is shown in \autoref{fig:results1}. Altering just one single
pixel of the input image (Attack A was found to reduce the
confidence for the actual classes by an average of about
16.2\% across our 38 test images. In Attack A, we were only
testing 56 pixels out of a total of 192 x 168 = 32,256 (about
0.2\% of all pixels) for each image, due to time constraints, so it
seems likely that if we were able to perform a more
comprehensive pixel search for each image, these individual
pixel-based attacks could be substantially more successful, as
they would be more likely to find a locally optimal perturbation.
However, it was interesting to see that Attack C, which tested
56 randomly chosen pixels and then another 56 randomly
chosen pixels on the result from the first perturbation, was only
marginally more successful than Attack B, which simply altered
the two best pixels from the first 56 chosen pixels, without
testing another 56. Both Attack B and Attack C produced
substantially greater decreases in confidence for the actual
class (on average, about 25.7\% and 28.7\%, respectively)
compared to Attack A. This seemed to offer tentative support
for a hypothesis that altering a greater number of pixels, while
being less particular about which ones to alter, could be a
more effective and much less computationally expensive
approach to generating attacks.

The attacks based on Approach 1 are time-expensive,
since after altering every pixel, the resulting image must be
classified and the change in confidence must be checked and only after iterating through each test pixel chosen from the
image can we select the best pixel to alter. Approach 1 rarely
resulted in misclassification, but it did produce a significant
average decrease in confidence that an image is its actual
class. This approach is not as realistic in real world
applications, since an attacker might not have access to the
confidence levels returned by the classifier for each class.
However, the approach is still useful for understanding how
robust the classifier is to small amounts of noise in specific
areas. We did not find any particularly consistent patterns in
terms of which regions of the faces caused the greatest
change in classifier confidence when a pixel from the region
was altered.

The attacks based on Approach 2, altering all pixels in the input image (i.e. Attacks D, E, and F), were substantially more successful in decreasing the confidence of the classifier that each image was its actual class, compared to the individual pixel-based Approach 1 attacks (i.e. Attacks A, B, and C). We
experimented with various iterations of Approach 2 attacks, using different magnitudes of perturbations. As expected, greater values for the magnitude of perturbation caused worse classifier performance, while they also resulted in images where the perturbations were more visibly apparent to a human, since the images became increasingly grainy and pixelated with increasing magnitudes of perturbation. Nevertheless, with the checkerboard approach, even with the perturbation magnitude range as high as 120-150 in Attack F, the perturbed images remained, while somewhat noticeably altered, still clearly recognizable to a human as the same person as the original image (see the example in \autoref{fig:attack_f_example})

The Dual Attack (Attack G), which first performed Attack D and then fed the perturbed image from that attack as input to Attack B, produced some improvement over Attack D alone, however it was not a particularly large improvement (72.0\% average decrease in classifier confidence for the actual class
for Attack F, compared to 65.8\% for Attack D alone), and was outperformed by Attack E, which simply increased the range of perturbation magnitudes to 60-90, compared to 30-60 in Attack D. Attack E achieved an average decrease in classifier confidence in the  actual class of 74.6\%, while the even more substantial perturbations of Attack F (120-150) achieved 84.2\%.

\begin{figure}[htbp!]
    \centering
  \includegraphics[width=0.7\linewidth]{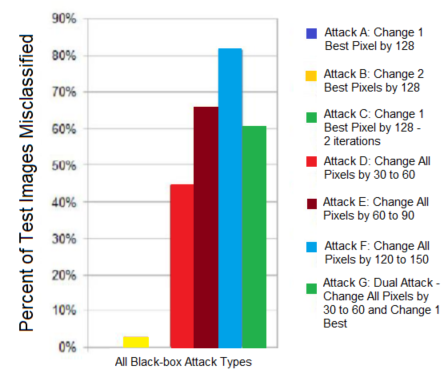}
  \caption{Graph of percent of images misclassified in the test dataset by the classifier ($PercentMisClass$) after each attack. Attack F results in the largest percent of misclassifications for the test images.}
  \label{fig:results2}
\end{figure}

\begin{figure}[htbp!]
    \centering
  \includegraphics[width=\linewidth]{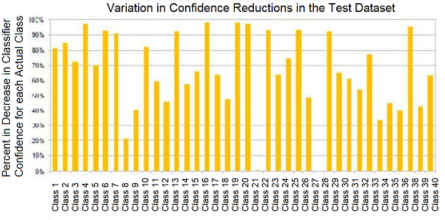}
  \caption{Graph showing variation in percent decreases in confidence for actual class across testing dataset after Attack D. One insight we found from these attacks is that larger surface area around eyes correlated (empirically) with adversarial success of the attacks. The reason why for some classes thee were little decreases can be attributed to the fact that those classes had smaller area around eyes and noise, which meant less exposure to the attacks.}
  \label{fig:variation_among_individiuals}
\end{figure}

In \autoref{fig:results2}, we present a bar graph of the percent of
images which were misclassified (PercentMisclass) for each
attack.
The Approach 1 attacks (A, B and C) rarely resulted in
misclassification of the input images. In the test runs for which
results are shown in 5, only one misclassification was achieved
through Attack B, and none for Attacks A or C.
The attacks which used Approach 2 (D, E and F), on the
other hand, were much more successful, resulting in
misclassification rates of 44.7\% for a 30-60 perturbation
magnitude range, 65.8\% for a 60-90 range, and 81.6\% for a
120-150 range. The Dual Attack (G) again fell between Attacks
D and E for this metric.

Another result from these attack was the variance between classes. A high degree of
variability in the level of success of the attacks was observed across different
individuals in our testing sample. \autoref{fig:variation_among_individiuals} demonstrates this in the case of the $ConfA$ results for Attack D. Each bar represents a different individual from the  dataset. While the attack caused almost half of the individuals to have the classifier’s confidence in their actual class decrease by 70\% or more, there were two individuals where the attack was almost entirely ineffective. 

As mentioned previously, we hypothesize
that the classifier’s robustness for these particular individuals
may have been because they represented minority groups
within our face database, and could have been in part due to
the training dataset not being very diverse.
This high degree of variability across individuals represents a possible weakness of our study. If we had time and
resources to train an image classifier on a larger and more
diverse facial image dataset, it seems possible that we may
have been able to produce more consistent results across
individuals in the testing dataset.
Greater time and computing resources could have also offered
us the ability to develop more comprehensive individual pixel-based attacks, testing greater numbers of pixels
per image, and altering more than two pixels per attack. More
experimentation in this area with larger datasets and analysis
of patterns in the optimal pixels selected could yield valuable
insights into whether there are certain regions of faces which,
when altered, are more likely to fool a facial recognition
classifier. Based on fairly anecdotal analysis of the pixels
chosen by our individual pixel-based attacks, we did not
observe any clear patterns in which pixels were chosen, but it
is possible that a more methodical approach, and perhaps
cluster analysis with larger datasets, could result in the
discovery of patterns that are not easily apparent to a human
observer when using the relatively small testing sample that we
used in our study.

A strength of our study is that we used iterative approaches to our attacks to attempt to find the minimal magnitudes of perturbation needed to achieve misclassifications.
While these iterative approaches would not necessarily be as
feasible when attacking a real classifier in the wild, the results
are useful for offering some tentative baselines for the amount
of perturbation that may typically be needed in order to fool a
facial recognition classifier. Additionally, the facial image
dataset that we trained our classifier on included some images
with very dark lighting conditions, and we used an advanced
convolutional DNN model fine-tuned for image recognition, so
we are fairly confident that the trained classifier that we
attacked was a fairly robust facial image classifier model.
Finally, we believe that our checkerboard approach to altering
all pixels in our attack images was quite successful, in that it
achieved high rates of misclassification while preventing the
resulting images from becoming unidentifiable to a human.

\section{Conclusion}
Through our experimentation with different adversarial attack
approaches, we found that altering all pixels in a face image by
a small amount (such that the image is still identifiable to a
human) was generally more effective than changing one or
two pixels by a large amount for causing a facial recognition
classifier to misclassify the image. Changing just one or two
strategically placed pixels reduced face classifier confidence
levels substantially, but typically did not result in
misclassifications. In our test, modifying all pixel values by a randomly chosen value between 30 and 60, in alternating
positive and negative directions, resulted in a 66\% decrease in
the classifier’s confidence for the actual class on average across
our 38 test images and a 45\% misclassification rate. When we
increased the level of perturbation to a range of 120 to 150, we
achieved an 84\% average decrease in the classifier’s confidence
for the actual class and an 82\% misclassification rate. The
resulting images from all of our attacks, while in some cases
visibly altered, were always still identifiable as the actual person
to a human. While our results produced lower misclassification
rates than some other studies that have used more sophisticated
white-box approaches with standard datasets, our results
demonstrate that black-box attacks can also be quite successful
in fooling image classifiers in the facial recognition domain.
\section{Future Work}
It would be useful to test our attacks on other facial
recognition classifier models to test for transferability. Just
because they were relatively successful against the Inception v3
Neural Network model that we trained would not necessarily
mean that they would be as successful against other models.
Models trained on datasets that have been noised using similar
approaches as our attacks, or models tuned specifically to be
more robust to adversarial attacks would likely be more difficult
to fool \cite{aab20driverfatigue}  \cite{mmstv17}. There have been adversarial studies in the past
that proved to make the models more robust after adding noise
and attacking the models under black-box and white-box adversarial
attack settings \cite{aab20audio} \cite{aakk20}. To test these attacks in the real world,
one could try to upload images altered by these attacks to social
media sites to see whether the sites’ classification algorithms
would be able to successfully and automatically tag them. A successful facial
recognition and classification could be easily translated to many
real life applications such as crime, bio-authentication, malware,
security, and spam detection, content tagging in social media
sites \cite{aplkc20}.
One could also apply our attacks to classifiers trained
on the MNIST or CIFAR standard image recognition datasets in
order to compare results with other studies that have used these
datasets, or trained on datasets from other domain areas, to test
for transferability to other categories of images. An additional
useful step would be to modify our attacks so that they can
handle color images in addition to grayscale ones. It would also
be beneficial to continue to experiment further with different
amounts of perturbation or designing functions to calculate the
most effective and minimized amounts of perturbation that
consistently result in misclassification. Finally, applying
individual pixel-based attacks to larger facial image datasets
could be a valuable project for exploring whether there are
certain regions of face images that tend to be particularly
influential for classifiers and which represent good regions to
target when developing adversarial attack algorithms.

\printbibliography[title=List of References]

\section{Appendix}
\subsection{Fast Gradient Sign Method}
\begin{figure}[htbp!]
\centering
\includegraphics[width=\linewidth]{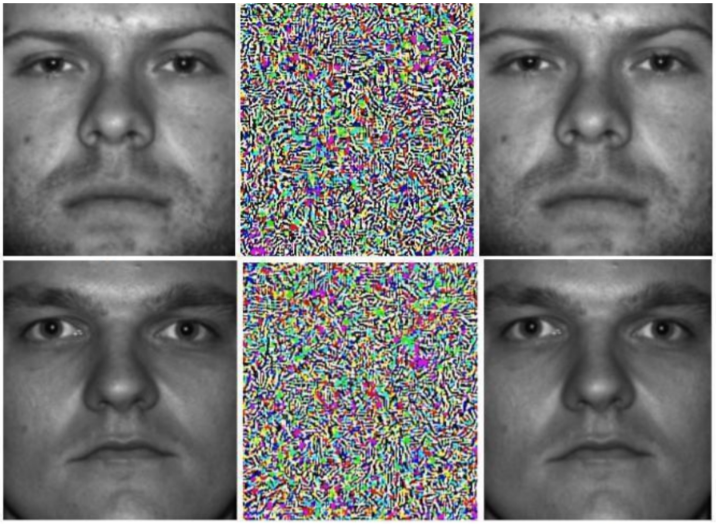}
  \caption{Sample visual representation when \emph{Fast Gradient Method} \cite{goodfellow2014explaining} is applied. In this sample, we attack a sample image from our dataset with \emph{Fast Gradient Sign Method} under a white-box attack setting. We add a perturbation that is invisible to human eye, yet the resulting image leads to a significant drop in the model’s confidence}
  \label{fig:fgsm}
\end{figure}

Fast Gradient Sign Method is an adaptation from current
literature, specifically from the studies of Papernot et al. \cite{gss15}
Figure \ref{fig:fgsm} shows the person 1 before the attack and the adversarial noise generated by the attack for that
specific person. Once the noise in Figure \ref{fig:fgsm} is generated, the
algorithm tries to find an epsilon value to multiply the noise
before adding to the original image. Resulting image is also shown on the right column in Figure \ref{fig:fgsm}. The
epsilon value is determined based on the breaking point where
the image is misclassified. Same procedures are also applied to the person 7 in Figure \ref{fig:fgsm}. Once the adversarial noise is added, the resulting image causes a misclassification for the facial DNN classifier. One important aspect to notice is that the resulting image looks almost the same as the first clean image. Such similarity between the adversarial image and the original image is in direct correlation with the strength of the attacking algorithm. Another aspect to notice is that, in FGSM, the adversary is assumed to have full knowledge regarding the classifier. Such knowledge includes the model architecture, confidence levels, parameters, etc. One can say that the strength of the algorithm comes from the abundance of such information possessed by the adversary. In real life, most of the time, the adversary does not have access to such parameters. The impracticality of the current proof-of-concept algorithms
was one of the main motivations behind this paper.

\begin{figure}[htbp!]%
    \centering
    \subfloat[Original Image Confidence: 0.625]{\includegraphics[width=0.4\linewidth]{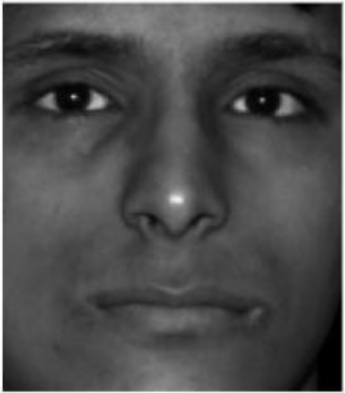} }%
    \qquad
    \subfloat[Attack C: Change 1 Best Pixel by 128 - 2 iterations Confidence: 0.444 ]{\includegraphics[width=0.4\linewidth]{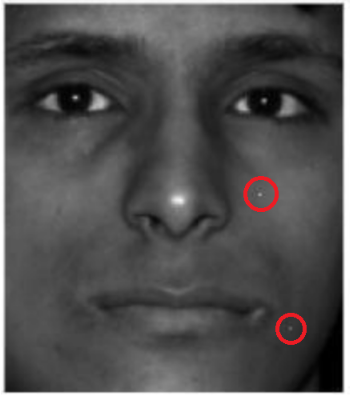} }%
    \\
    \centering
    \subfloat[Attack D: Change All Pixels by 30 to 60 pixels. Confidence: 0.344]{\includegraphics[width=0.4\linewidth]{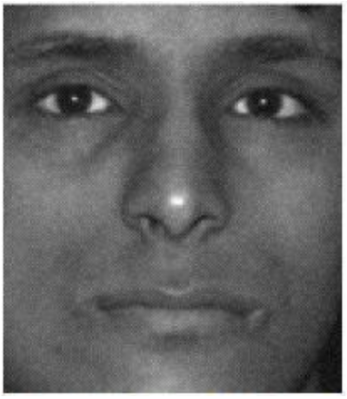} }%
    \qquad
    \subfloat[Attack F: Change All Pixels by 120 to 150. Confidence: 0.0423 (Misclassified)]{\includegraphics[width=0.4\linewidth]{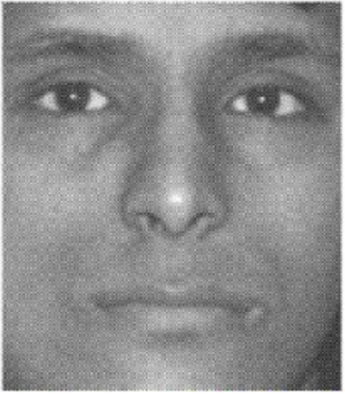} }%
    \caption{The original image of \emph{person 35} and resulting images from three of the attacks we developed, along with the confidence levels that the classifier returned for each of those images. Two pixels are changed for subplot (b) as they can be seen via red circles.}%
    \label{fig:person35}
\end{figure}

\end{document}